\pdfoutput=1

\documentclass[11pt]{article}

\usepackage{emnlp2021}

\usepackage{times}
\usepackage{latexsym}

\usepackage[T1]{fontenc}

\usepackage[utf8]{inputenc}

\usepackage{microtype}

\usepackage{amsmath}
\usepackage{amssymb}

\usepackage{graphicx} 

\title{Path-Enhanced Multi-Relational Question Answering with Knowledge Graph Embeddings}

 \author{Guanglin Niu, Yang Li, Chengguang Tang, Zhongkai Hu, Shibin Yang,\\ \bf{Peng Li, Chengyu Wang, Hao Wang, Jian Sun}\\
        Alibaba Group}

\begin{document}
\maketitle
\begin{abstract}
 The multi-relational Knowledge Base Question Answering (KBQA) system performs multi-hop reasoning over the knowledge graph (KG) to achieve the answer. Recent approaches attempt to introduce the knowledge graph embedding (KGE) technique to handle the KG incompleteness but only consider the triple facts and neglect the significant semantic correlation between paths and multi-relational questions. In this paper, we propose a \underline{\textbf{P}}ath and \underline{\textbf{K}}nowledge \underline{\textbf{E}}mbedding-\underline{\textbf{E}}nhanced multi-relational \underline{\textbf{Q}}uestion \underline{\textbf{A}}nswering model (PKEEQA), which leverages multi-hop paths between entities in the KG to evaluate the ambipolar correlation between a path embedding and a multi-relational question embedding via a customizable path representation mechanism, benefiting for achieving more accurate answers from the perspective of both the triple facts and the extra paths. Experimental results illustrate that PKEEQA improves KBQA models' performance for multi-relational question answering with explainability to some extent derived from paths.
\end{abstract}

\section{Introduction}

Knowledge graphs store rich facts in a multi-relational graph structure such as Wikidata \cite{wikidata} and Freebase \cite{BGF:Freebase}. One of the popular applications based on KGs is Knowledge Base Question Answering (KBQA). Multi-relational KBQA requires multi-hop reasoning in the KG corresponding to the question for obtaining the correct answer \cite{SRN, KVMem-EMNLP, InterRN}.

However, the existing KGs are inevitably incomplete. The missing links of the answer entities lead to an out-of-reach issue. To address this issue, two streams of studies are proposed: (1) exploiting extra text corpus:  both GRAFT-Net \cite{KBText} and PullNet \cite{PullNet} incorporate the KG and the entity-linked text to build a question-specific sub-graph for extracting answers from this sub-graph containing the text and KG. (2) Exploiting KG embeddings: KEQA \cite{KEQA} leverages the embeddings of the question’s topic entity, the relation, and the tail entity to predict the answer entity via the score function inspired by KGE approaches. EmbedKGQA \cite{EmbedKGQA} expends KEQA for multi-relational KBQA by calculating the answer score from the leaned question embedding and the entity embeddings. However, the valuable text corpus is not always available, limiting the scalability of the text-enhanced models. On the other hand, the existing KGE-based KBQA models ignore the significant path information to evaluate the correlation between paths and multi-relational questions. Besides, EmbedKGQA cannot make full use of the relation embeddings.

To address the above challenges, we develop a path and knowledge embedding-enhanced multi-relational KBQA model (PKEEQA) to employ the semantics of paths and the generalization of KG embeddings. Notably, the PKEEQA model is a pluggable module that works in the inference stage without retraining. It could provide a supplementary semantic correlation between the multi-hop paths and the multi-relational questions via a customizable path representation mechanism for improving the accuracy of the answer. More interestingly, the paths can also provide explainability on how the answer entity is reached from the topic entity. The experiments are conducted to demonstrate the effectiveness of our model.

The contributions of our work are: (1) we are the first to incorporate path information with KG embeddings for multi-relational KBQA. (2) We develop a customizable path representation mechanism for three types of KGE models. The ambipolar correlation between path and question embeddings is obtained to seek accurate answers. (3) The experimental results illustrate the effectiveness of our model incorporating paths and KGE.

\section{Related Work}

\subsection{Multi-relational KBQA}

\label{multi-relational KBQA}
Compared with simple KBQA, multi-relational KBQA is a more challenging task, and our study focuses on it. Specifically for multi-relational KBQA,  KVMem \cite{KVMem-EMNLP} stores KG triples in a key-value structure and reasoning on the stored key-value memories to predict the answer. BAMnet \cite{BAMNet} modifies KVMem as a bidirectional attentive memory network model. SRN \cite{SRN} formulates multi-relational KBQA task as a Markov decision problem and introduces reinforcement learning to search the answer in the KG. GRAFT-Net \cite{KBText} and PullNet \cite{PullNet} both construct a question-specific sub-graph by combining KG and text corpus, where the text corpus can enrich the KG to alleviate the KG sparsity. Whereas, the adequate text corpus is not always available. Inspired by the KG embedding technique for KG completion, EmbedKGQA \cite{EmbedKGQA} employs the pre-trained entity embeddings by KGE and the question embedding to calculate the score of each candidate answer entity.

\subsection{KGE Models}
\label{KGE}
In this paper, we classify the KGE models into three types according to the characteristic of the score function, namely: (1) \textbf{Additive} operator-based models. TransE \cite{Bordes:TransE} regards a relation as the translation operation between an entity pair. The vectors of entities and relations are expected to satisfy $\textbf{h}+\textbf{r}\approx\textbf{t}$ when the triple $(h,r,t)$ holds. And then, some extension models of TransE are developed, including TransR \cite{Lin-a:TransR}, TransF \cite{TransF} and TransD \cite{TransD}. (2) \textbf{Multiplicative} operator-based models. Based on the idea of tensor decomposition in RESCAL \cite{RESCAL}, the value of each triple can be calculated by the product of the head and the tail entity vectors $\textbf{h},\textbf{t}$ and the relation matrix $\textbf{M}_r$, expecting $\textbf{h}^\top\textbf{M}_r\textbf{t}=1$. DistMult \cite{DistMult}, \cite{HolE} and ComplEx \cite{Trouillon:ComplEx} all simplified RESCAL. (3) \textbf{Hardmard product} operator-based models. RotatE \cite{RotatE} employs Hardmard product to treat the relation as a rotation operation between an entity pair, which expects $\textbf{h} \circ \textbf{r} \approx \textbf{t}$.

\begin{figure*}
  \centering
  \includegraphics[scale=0.75]{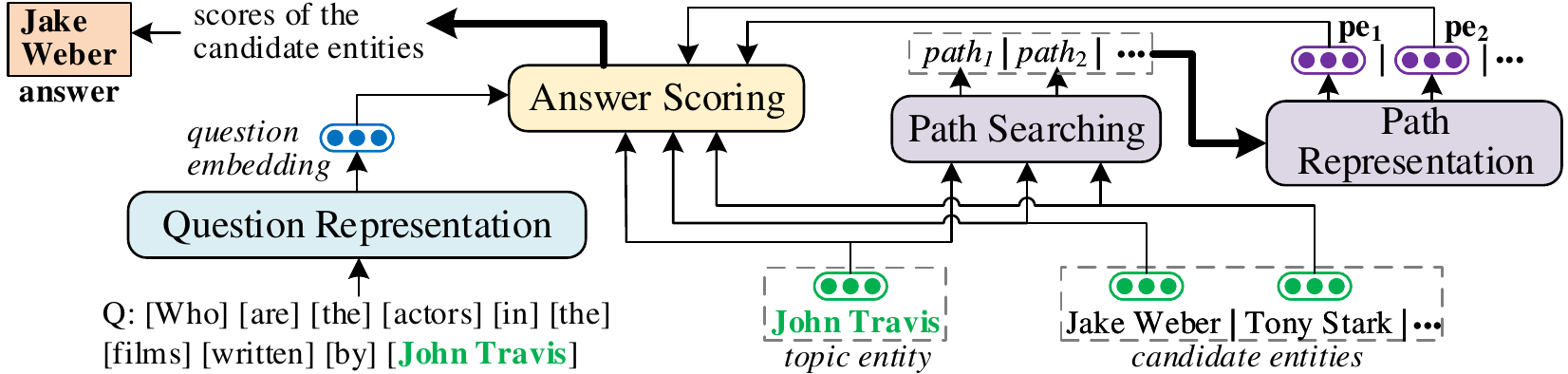}
  \caption{The overall architecture of our model PKEEQA. $path_1$ and $path_2$ are the relation paths from the topic entity $John\ Travis$ to the candidate entities $Jake\ Weber$ and $Tony\ Stark$, respectively. $\textbf{pe}_1$ and $\textbf{pe}_2$ corresponds to the path embeddings of $path_1$ and $path_2$.}
  \label{figure1}
\end{figure*}

\section{Methodology}

Inspired by the idea of introducing KGE into multi-relational KBQA, we firstly pre-train KG embeddings via any KGE model and train the KBQA model by the score from the triple containing the question embedding learned by LSTM and two pre-trained entity embeddings, which is calculated in consistence with the KGE approach. Then, in the inference process, the shortest relation path from the topic entity to each candidate answer entity is extracted if the path exists. A customizable path representation mechanism can learn the path embedding. Furthermore, the score of each candidate entity consists of both the triple consistency and the ambipolar correlation derived from the path and question embeddings. Finally, we obtain the answer entity via the ranking of the candidate answering scores. The overall architecture of our model PKEEQA is shown in Figure \ref{figure1}.

\subsection{Customizable Path Representation}

 The existing path encoder with the architecture of LSTM in some commonsense QA models such as KagNet \cite{KagNet} is not available for our task due to extremely high complexity caused by a large number of paths. Thus, we propose a simplified and customizable path representation mechanism corresponding to various KGE models. In terms of the three types of KGE models described in section \ref{KGE}, the customizable path representation mechanism is designed as the following three forms in accordance with the KGE model:

(a) For \textbf{additive} operator KGE model (i.e., TransE\cite{Bordes:TransE}):
\begin{equation}
  \textbf{p} = \sum_{r_i\in P(h,a)}\textbf{r}_i \label{eq1} \\
\end{equation}
where $P(h,a)$ denotes a set containing all the relations in the path sequence from the topic entity $h$ to the candidate entity $t$. $\textbf{r}_i$ indicates the embedding of the $i$-th relation $r_i$ in the set $P(h,a)$. $\textbf{p}$ is the path embedding in the relation embedding space.

(b) For \textbf{multiplicative} operator KGE model (i.e., ComplEx\cite{Trouillon:ComplEx}):
\begin{equation}
  \textbf{p} = \prod_{r_i\in P(h,a)}\textbf{r}_i \label{eq2} 
\end{equation}

(c) For \textbf{Hadamard productive} operator KGE model (i.e., RotatE\cite{RotatE}):
\begin{equation}
  \textbf{p} = \textbf{r}_1 \circ \textbf{r}_2 \circ \cdots \circ \textbf{r}_n \label{eq3}
\end{equation}
where $P(h,t)$ denotes the hardmard product. $n$ is the total number of the relations in the path.

\subsection{Path and Embedding-Enhanced Answer Selection}

In the inference process, the score function is applied for evaluating the probability of each candidate answer. Similar to the three types of path representation mechanism, the score function also processes three forms for various KGE models:

(a) For \textbf{additive} operator KGE model:
\begin{equation}
  \textbf{E} = -\| \textbf{h} + \textbf{q} - \textbf{a} \| + \alpha \cdot \tanh(\rm{sim}(\textbf{q}, \textbf{p}))\label{eq4} \\
\end{equation}
where $\textbf{h}$, $\textbf{q}$ and \textbf{a} are embeddings of the topic entity embedding, the question and the candidate answer entity. $\rm{sim}(\textbf{q}, \textbf{p})$ means the similarity between the question embedding and the path embedding, in which we select cosine distance function to calculate the similarity. $\alpha$ denotes the weight for the trade-off between the triple and the path-question correlation. In specific, the output value of $\tanh$ function is in the range of [-1,1], it could reflect the ambipolar correlation between the path and the question because the paths mismatching the question even deduce the less likely answers. The score specific to the path implies the positive or negative correlation between a path and the given question, which can be further combined with the triple scoring for exploring the more accurate answer.

(b) For \textbf{multiplicative} operator KGE model:
\begin{equation}
  \textbf{E} = \textbf{E}_t + \alpha \cdot \tanh(\rm{sim}(\textbf{q}, \textbf{p}))\label{eq5} \\
\end{equation}
where $\textbf{E}_t$ is the score function specific to the triple:
\vspace{-0.3cm}
\begin{equation}
\textbf{E}_t=
\begin{cases}
\sum_{d=1}^{k}{h^d q^d a^d}& \text{ $in\ real\ space$ } \\
Re(\sum_{d=1}^{k}{h^d q^d \bar{a}^d})& \text{ $in\ complex\ space$ }
\end{cases}\label{eq6}
\end{equation}

(c) For \textbf{Hadamard productive} operator KGE model:
\begin{equation}
  \textbf{E} = -\| \textbf{h} \circ \textbf{q} - \textbf{a} \| + \alpha\cdot\tanh(\rm{sim}(\textbf{q}, \textbf{p}))\label{eq7} \\
\end{equation}

According to Eqs. \ref{eq4} - \ref{eq7}, we can rank the scores corresponding to all the candidate answer entities. Then, the candidate answer entity with the highest score is selected as the predicted answer.

\begin{table}
\caption{Statistics of the experimental datasets.}
\centering
\begin{tabular}{c|ccc}
\hline
Dataset		      & \#Train	    & \#Test	& \#Valid    \\
\hline
 MetaQA (1-hop)   & 96,106	    & 9,947     & 9,992    \\
 MetaQA (2-hop)   & 118,948	    & 14,872    & 14,872    \\
 MetaQA (3-hop)   & 114,196	    & 14,274    & 14,274    \\
\hline
\end{tabular}
\label{table1}
\end{table}

\section{Experiments}

\subsection{Dataset and Experimental Settings}

MetaQA is a widely-used multi-relational KBQA benchmark dataset and contains 1-hop, 2-hop, and 3-hop questions, respectively. It consists of more than 400K questions and a KG with 43K entities, nine relations, and 135K triples. In particular, we generate the inverse relations in the KG to ensure all the relations in each path from the topic entity to the candidate answer entity are in the same direction. Therefore, the amount of relations is doubled. The statistics of MetaQA are provided in Table \ref{table1}.

We compare PKEEQA with the state-of-the-art multi-relational KBQA baselines, including VRN \cite{VRN}, KVMem \cite{KVMem-EMNLP}, GraftNet \cite{KBText}, PullNet \cite{PullNet} and EmbedKGQA \cite{EmbedKGQA}.

The dimension of the pre-trained entity and relation embeddings is set to 200. We tune all the hyper-parameters via grid search strategy. The optimal hyper-parameters are selected as: the hidden states of LSTM for learning the question embedding is 256, the weight $\alpha$ is 0.1, the batch size is set as 128, and the learning rate is 0.0005. The experiments are conducted on Ubuntu 16.04 with 64GB i9-9900X CPU and one Nvidia 2080Ti GPU.

\begin{table}[!t]\small
\renewcommand{\arraystretch}{1.2}
\setlength{\tabcolsep}{1mm}
\caption{Multi-relational KBQA results (\%) on MetaQA with the settings of the full KG (KG-Full) and half of the KG (KG-Half). The best results are in \textbf{bold}, and the second-best results are \underline{underlined}.}
\centering
\begin{tabular}{c|ccc|ccc}
\hline
                        & \multicolumn{3}{c}{KG-Full}       & \multicolumn{3}{|c}{KG-Half}	\\
   
    \textbf{Model}      & 1-hop     & 2-hop     & 3-hop     & 1-hop     & 2-hop     & 3-hop \\
\hline
VRN             & 97.5      & 89.9      & 62.5      & -         & -         & -   \\
GraftNet        & 97.0      & 94.9      & 77.7      & 64.0      & 52.6      & 59.2   \\
PullNet         & 97.0      & \textbf{99.9}      & 91.4      & 65.1      & 52.1      & 59.7   \\
KV-Mem          & 86.2      & 82.7      & 48.9      & 63.6      & 41.8      & 37.6   \\
EmbedKGQA       & \underline{97.5}      & 98.8      & \underline{94.8}      & \underline{83.9}      & \underline{91.8}      & \underline{70.3}   \\
\hline
PKEEQA                  & \textbf{98.0} & \underline{99.3}     & \textbf{95.1}     & \textbf{85.0} & \textbf{92.5}     & \textbf{70.7}  \\
\hline
\end{tabular}
\label{table2}
\end{table}

\begin{table}\small
\renewcommand{\arraystretch}{1.2}
\caption{Ablation study results (\%) of multi-relational KBQA on MetaQA. 1-H, 2-H and 3-H indicate 1-hop, 2-hop and 3-hop questions, respectively.}
\centering
\begin{tabular}{c|ccc|ccc}
\hline
                        & \multicolumn{3}{c}{KG-Full}       & \multicolumn{3}{|c}{KG-Half}	\\
    \textbf{Model}      & 1-H     & 2-H     & 3-H     & 1-H     & 2-H     & 3-H \\
\hline
Whole          & \textbf{98.0} & \textbf{99.3}     & \textbf{95.1}     & \textbf{85.0} & \textbf{92.5}     & \textbf{70.7}  \\
\hline
-Path                   & 97.6      & 98.9      & 94.7      & 83.9      & 92.0      & 70.3   \\
-PT+PS                  & 97.2      & 99.0      & 94.6      & 83.1      & 92.1      & 70.5   \\
\hline
\end{tabular}
\label{table3}
\end{table}

\subsection{Experimental Results}

On average, PKEEQA spends 10h for training and 16h for inference. On MetaQA dataset with full KG, our model consistently outperforms other baselines on 1-hop and 3-hop questions, and achieves comparable results on 2-hop questions as shown in Table \ref{table2}. PKEEQA performs better than almost all the baselines of 2-hop questions and all the baselines of 3-hop questions benefited from leveraging paths for supplementing the correlation between paths and multi-relational questions. Besides, our model PKEEQA obtains better performance on 1-hop compared to the baselines because the additional path information provides more semantic association between the topic entity and the answer entity, improving the accuracy of the answer. 

We also evaluate the performance of our model on MetaQA with half of the KG (MetaQA-half). As shown in Table \ref{table2}, PKEEQA obtains the best performance than other baselines. Compared to the baseline models without KG embeddings, PKEEQA and EmbedKGQA both take advantage of KG embeddings for KBQA on the sparse KG with the ability of KG completion. In addition, PKEEQA outperforms EmbedKGQA for exploiting paths to enrich the links between entities to address the sparseness of the KG.

Particularly, we have compared three types of KGE models for pre-training TransE, ComplEx and RotatE and the corresponding path representation.s PKEEQA based on ComplEx and multiplicative operator path representation performs better than the other two types. We also implement KG completion of the three types of KGE models and ComplEx outperforms the others, which verifies the best performance of PKEEQA based on ComplEx and multiplicative operator path representation. The results in Table \ref{table2} and Table \ref{table3} are based on ComplEx.

\subsection{Ablation Study}

To investigate the effectiveness of introducing path information into multi-relational KBQA with KG embeddings, we conduct ablation studies by removing path information from the whole model (-Path) and just replacing tanh with sigmoid function for the score specific to paths (-PT+PS). The results of the ablation study are shown in Table \ref{table3}. We can observe that our whole model outperforms the other two ablated models. It illustrates the path information (from -Path) and the ambipolar correlation of path-question pair (from -PT+PS) both play pivotal roles in multi-relational KBQA.

\subsection{Case Study}

\begin{figure}
  \centering
  \includegraphics[scale=0.9]{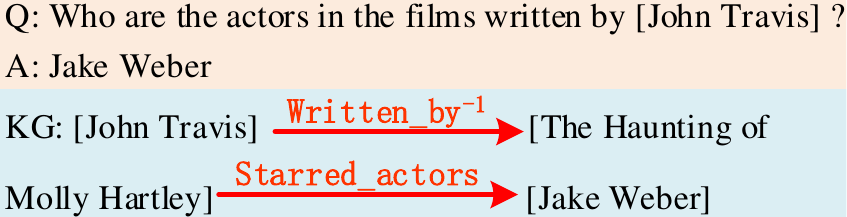}
  \caption{An example of the explainable KBQA on MetaQA (2-hop). The superscript "-1" means the inverse version of the relation.}
  \label{figure2}
\end{figure}

As shown in Figure \ref{figure2}, given a 2-hop question on MetaQA, our approach could not only output the correct answer but also provide the path from the topic entity to the answer entity. This path is essential for explaining the process of obtaining the answer from the KG and increasing the user’s credibility with the answer.

\section{Conclusion}

In this paper, we propose a novel model PKEEQA via incorporating path information with KG embeddings for multi-relational KBQA. A customizable path representation strategy is developed for any pre-trained KGE model. And then, we could evaluate the ambipolar correlation between the multi-hop paths and the multi-relational questions to seek the accurate answer via the designed answer scoring. The experimental results illustrate the effectiveness of our model that introduces paths into candidate answer scores to improve the accuracy of selecting the correct answer.

\bibliography{anthology}
\bibliographystyle{acl_natbib}

\end{document}